\documentclass{amsart}
\usepackage{cite}
\usepackage{hyperref}
\usepackage{bm} 
\usepackage{booktabs}
\usepackage{graphicx}
\usepackage{float}

\author{Attila Egri-Nagy$^1$, Antti T\"orm\"anen$^2$}
\address{$^1$Akita International University\\Department of Mathematics and Natural Sciences\\ Yuwa, Akita-City 010-1292, Japan}
\address{$^2$Nihon Ki-in -- Japan Go Association\\7-2 Gobancho, Chiyoda City,\\ Tokyo 102-0076, Japan}
\email{egri-nagy@aiu.ac.jp,\ tormanen.antti@gmail.com}
\title[Derived metrics for the game of Go]{Derived metrics for the game of Go -- intrinsic network strength assessment and cheat-detection}

\begin{document}
\maketitle
\begin{abstract}
 The widespread availability of superhuman AI engines is changing how we play the ancient game of Go.
The open-source software packages developed after the AlphaGo series shifted focus from producing strong playing entities to providing tools for analyzing games.
Here we describe two ways of how the innovations of the second generation engines (e.g.~score estimates, variable komi) can be used for defining new metrics that help deepen our understanding of the game.
First, we study how much information the search component contributes in addition to the raw neural network policy output. This gives an intrinsic strength measurement for the neural network. Second, we define the effect of a move by the difference in score estimates. This gives a fine-grained, move-by-move  performance evaluation of a player. We use this in combating the new challenge of detecting online cheating.
\end{abstract}

\section{Introduction}
The game of Go is an ancient board game with simple rules and enormous complexity.
It was the last grand challenge for artificial intelligence (AI) in abstract board games, when the challenge is understood as beating the best human player, not as solving the game.

AlphaGo (AG)~\cite{AlphaGo2016} made history by being the first superhuman Go AI engine.
By using deep neural networks, AG had a way to integrate expertise of master players, further enhanced by reinforcement learning and self-plays.
AlphaGo Zero (AGZ)~\cite{AlphaGoZero2017} improved the results by removing human expertise from the training process.
These developments are revolutionary in AI.
However, the real revolution came afterwards, when the technology became available to all players.

Several new implementations followed the success of AG and AGZ~\cite{ELF2019, SAI7x7-2019, SAI9x9-2019, LZ, KataGo2019}.
Given some computational resources, now anyone can build a deep learning Go engine \cite{pumperla2019deep}.
Moreover, just a standard gaming PC is capable of providing superhuman play and analysis.

The new implementations did not just recreate the same architecture, but several of them went beyond it in terms of providing more information about the game.
We call second-generation engines those that can play with variable komi (the points White get in the beginning as a compensation for not making the first move) and give information about the expected score, not just the probability of winning.
This fixes the problem of ‘slack’ moves, which AG was famous for.
These were interpreted as mistakes first, but then it was realized that once a win is secured, the neural network has no preference for choosing efficient moves.

After AG, the focus shifted from creating a superhuman Go-playing entity to developing tools that help in understanding the game and in the learning process of human players. Now, the main usage of superhuman AIs is game analysis.

\subsubsection*{The structure of the paper} First we will review the basic measures used in deep learning Go AIs, followed by the suggested derived measures. Then we will describe two applications, one for measuring network strength intrinsically, and one for online cheat detection. Next, we describe the developed software tool and close the paper with discussion.

\section{Basic Measures}

The AGZ-like systems are based on deep reinforcement learning.
Therefore we can describe their functioning in games and in analysis (we are not considering training here) in terms of neural networks and Monte-Carlo tree searches. Here we describe three important measures: the visit count, the win rate, and the score mean.

We denote a state of the game (the board position) by $s$, specifying the turn number as an index when needed. This way, $s_0$ denotes the empty board.
We denote a move (action) on turn $i$ by $a_i$: this takes the board position $s_{i-1}$ to $s_i$. In particular,  the first move $a_1$ takes $s_0$ to $s_1$. The action can be a pass.
\subsection{Visit count: $N(s,a)$}

For move $a$ at board position $s$ the visit count $N(s,a)$ is the number of times the search algorithm examined a variation starting with $a$.
The Monte-Carlo tree search methods keep track of how many times a node in the search tree gets visited. In AGZ \cite{AlphaGoZero2017}, the move selection is solely based on the visit count, since the search algorithm keeps visiting the promising moves.
Roughly speaking, the number of visits measures how many times a particular candidate move is considered, how `interesting' it is.
Another way to look at the visit count is to use it as a reliability measure. A move may look very promising with a high chance of winning, but with just a few visits we cannot trust its value. While analysis GUIs expose this value, it may be less used by the end users.

\subsection{Value function: $V(s)$, win rate}

The value function $V(s)$ gives the probability of winning the game at a board position $s$.
For the sake of simplicity, unless otherwise stated we consider the value function from the perspective of Black.

In AG \cite{AlphaGo2016}, a dedicated network was trained for estimating the value function. In AGZ \cite{AlphaGoZero2017} it became another head of the same network shared by the policy head.
It was realized that the same neural computation can be used both for predicting moves and for deciding who is winning.

\subsection{Score mean: $\mu_s$}

Convolutional neural networks can have different heads, giving other values beyond a probability distribution for the next move. They can be trained to predict the score lead, the score difference at the end of the game \cite{KataGo2019,SAI7x7-2019,SAI9x9-2019}. The score value head combined with the Monte-Carlo search methods give statistical information about the outcome of the game: the \emph{score mean} value. It can be interpreted as the estimated score difference between the players at the end of the game.

\emph{How reliable is the score mean?} It is part of the loss function for the neural network's training \cite{KataGo2019}, therefore the reliability of the estimate should increase with the strength of the network. Searching for  an indicator, we tried several handcrafted self-plays with KataGo's final 40 blocks network, starting the game with a balanced integer komi. Handcrafted means that the move is selected by a human operator after extensive analysis, to make sure that the choice is the best possible by the network with no time control. These games reliably end up in draws, indicating the stability of $\mu_s$.

There is an analogy for score mean in chess, where the advantage is measured by centipawns ($\frac{1}{100}$th of the value of the pawn). Score mean has a similar role in Go, with the added benefit that it fully captures the goal of the game. In chess one may need to consider distance from checkmates as well.

\subsection{Score mean vs.~win rate – the human perspective}

The relationship between score mean and win rate is not straightforward.
On average, positive score mean comes with higher than 50\% winning chance, but this is not a strict rule.
There can be a situation in which there is a high chance of loosing by a small margin, but there is still the possibility with low probability of capturing a big group.
Therefore, positive mean score can be associated with less than 50\% win rate.
Further properties of the connection can be demonstrated with two simple examples: a high-handicap game and a general consideration of the dynamics of score mean throughout a game.

In a high-handicap game, Black's advantage might be eroding steadily (reflected in the gradual decline of score mean), while the win rate stays flat above 90\%. Then, suddenly the win rate switches when Black's score mean becomes negative. Analysing this situation without the score mean could mislead us to search for a special meaning for the last little mistake, while it is just one of many.

Every move played in a game reduces the number of its future possibilities. As a game proceeds, its score estimate becomes more likely to be realized; so, while a game's score mean might remain constant and close to even, the game's win rate will eventually drift to an extreme.

Therefore, while win rate is a useful measure for the AI, it is often unintuitive for human players and it can be misleading. A relatively small mistake can cause a big shift in win rate. This effect is further amplified if a game is nearing its end.

Score mean is a useful measure for human players for two reasons. Firstly, strong human players themselves tend to estimate the values of moves in points, so the score mean values can be easily understood. Secondly, unlike the win rate, the score mean is not affected by the stage of the game. For example, a move that loses one point in terms of the score might cause a win rate shift of 50\% in the late game, but only 5\% in the early game. A human player cannot visualise this win rate shift, but the one-point loss is easy to understand.

\section{Derived Measures}

Based on the inner measures of deep learning Go AI engines, we define new measures to increase their usability and explainability. These can be viewed as new perspectives, from which we can understand the games and their analyses better.

\subsection{The effect of a move: $\delta(a)$}

The effect $\delta(a)$ is the difference between the score mean after and before a move $a$: $\delta(a)=\mu_{s_{i+1}}-\mu_{s_i}$, when $a$ takes board position $s_i$ to $s_{i+1}$.
The difference in the corresponding win rates was used first in Go GUIs, but as discussed before, the score mean is more stable and more informative, thus they quickly included the effect as well.

By gathering statistical information of the effects throughout a game (average of the effects, deviations from the mean, cumulative moving average of the effects) we can characterize the playing skill of a player. However, this alone cannot give a rating to a player, as the effects also depend on the type of the game played.

\subsection{Search gaps: hit rate and KL-divergence}

\begin{table}[t]
\begin{center}
  \begin{tabular}{lllll}
    \toprule
    & \multicolumn{4}{c}{Networks (256 channels)}\\
& \multicolumn{2}{c}{20 blocks} & \multicolumn{2}{c}{40 blocks}
\\\cmidrule(lr){2-3}\cmidrule(lr){4-5}

    Games & early & final & early & final \\
\midrule
1846 ``Ear-reddening'' & 61.04\% &61.96\% & 60.43\% & 64.11\% \\
  325 positions  & 199  & 202  & 197 & 209 \\
\midrule
  2016 ``Move 37'' &  55.66\% & 55.66\%  & 56.60\% & 58.01\% \\
212 positions   & 118 & 118 & 120 & 123\\
\midrule
  2019 Meijin & 53.25\%   & 58.44\%  & 53.68\%  & 61.03\% \\
231 positions  & 123 & 135 & 124 & 141\\
\midrule
    2020 kyu game & 58.51\% & 57.44\%  & 63.83\% & 61.17\%   \\
 188 positions & 110 & 108 & 120 & 115\\
\bottomrule
  \end{tabular}
\end{center}
\caption{Comparison of hit rate percentages of different networks.}
\label{HitRates}
\end{table}

\begin{figure}[t]
\centering \includegraphics[width=0.4\textwidth]{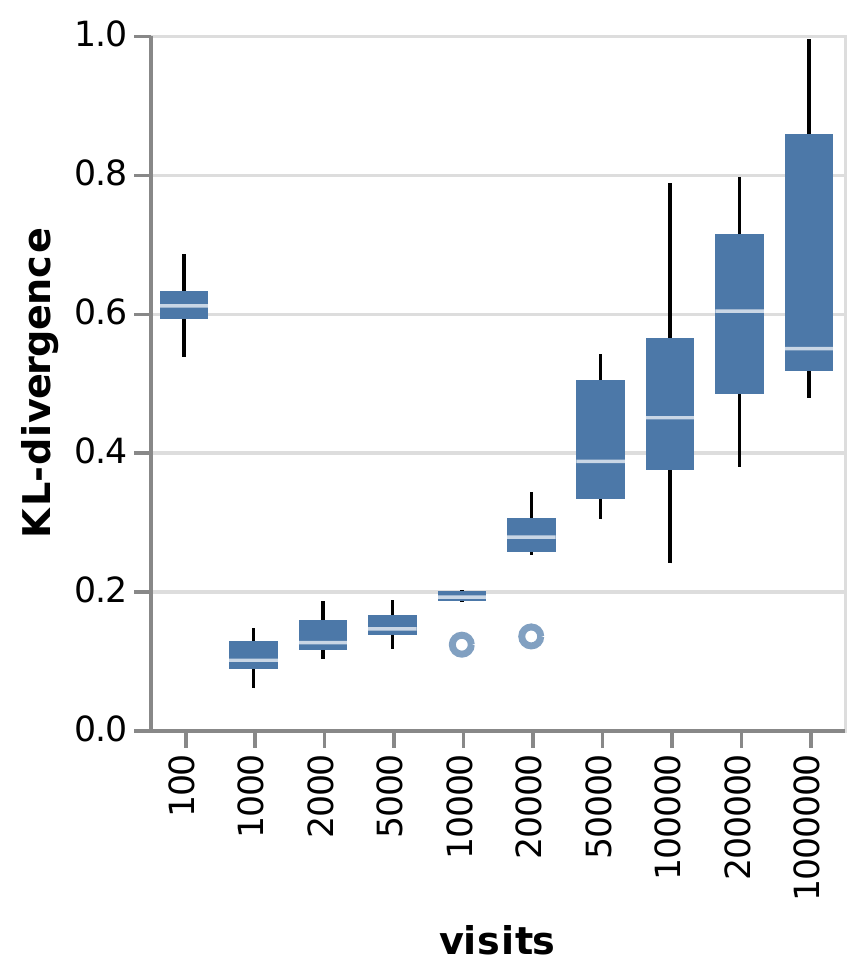}
\caption{Analyzing the turn of move 37 in the second game of the AlphaGo-Lee Sedol match. The same turn is analyzed with different visit counts, 7 different runs for each visit count.}
\label{calibration}
\end{figure}

$P(s,a)$, the prior probability of move $a$ at board position $s$, is provided by the raw network output. This probability distribution $\mathbf{p}$ is called the \emph{policy}.
The tree search guided by this policy then produces an updated policy $\bm{\pi}$, the probability distribution of good moves after search. $\bm{\pi}$ can be simply defined by the visit counts \cite{AlphaGoZero2017, pumperla2019deep}, as it is a good measure of the value of a candidate moves, given that enough simulations were made.
The disparity between $\bm{p}$ and $\bm{\pi}$ is the \emph{search gap}, which can be measured in different ways.

\subsubsection{Hit Rate} \emph{How many times does the search select the same move as the top move in the raw policy?} Clearly, this depends on the length of the search. If we just allow a couple of simulations, then this number will be high. So hit rate is relative to number of simulations.

\subsubsection{KL-divergence}
The Kullback-Leibler divergence\cite{KL1951} is a fundamental tool for comparing two discrete probability distributions, $P$ and $Q$.
$$D_{KL}(P\parallel Q)= \sum P(x)\ln \frac{P(x)}{Q(x)}$$
It is a measure of the disparity of the two distribution, although it is not a distance metric.
It measures how much information we gain if we use the distribution $Q$ instead of $P$.
It is a positive number, and it is zero when the distributions are the same.
The KL-divergence is a natural choice in the context of deep learning, since the closely related cross-entropy is used as a loss function for training the networks \cite{pumperla2019deep}.

We want to measure $D_{KL}(\bm{p}\parallel\bm{\pi})$, but there are a couple of issues. Both $\bm{p}$ and $\bm{\pi}$ can have zero entries. There are illegal moves (a stone is already there, suicide move, or a ko situation), and the search will also visit only a subset of the possible moves, so in general we do not have visit counts for all legal moves. Therefore, we take the actually visited moves in the search tree, and define $\bm{\pi'}$ by their visit counts and using normalization. So $\bm{\pi'}$ is the probability distribution of the moves considered by the network.  Note that this is now well-defined, while we used $\bm{\pi}$ informally before. Then we take the set of moves included in $\bm{\pi'}$ and find the corresponding probabilities in $\bm{\pi}$, and restricting to those moves, we normalize and get $\bm{p'}$.

As a rough but useful analogy, we can say that the output of the neural network corresponds to human intuition, while the search algorithm resembles step-by-step logical thinking. Just as humans mix these two types of thinking, the computer combines the deep neural networks with tree search. We want to measure the strength of intuition of the deep neural networks. This can be done by comparing the policy with or without tree search.

\section{Application: Intrinsic Strength of Networks}

\begin{figure}[t]
\centering \includegraphics[width=0.5\textwidth]{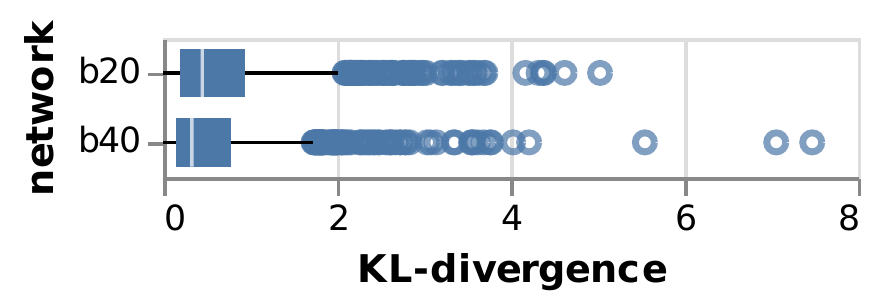}
\caption{Measuring the KL-divergence after 100,000 visits for a randomly chosen position in 915 strong amateur games.}
\label{KGS}
\end{figure}

\emph{How far are the deep neural networks from perfect minimax play?} For now, the universally agreed answer to this almost philosophical question is that they are very far.
We stop  training a network due to external reasons (e.g.~the cost of computational resources), not because we reached a theoretical limit for improvement. If a network played perfectly, the reported win rates could be more polarized, tending to one of the values 0, 0.5, and 1.0. Also, in that case, we would not need the tree search.

\emph{How long shall we run a game analysis?} This is a more practical, but related question. Can we simply use the raw network output policy? As a calibration test, we analyzed a game position with different visit counts. Since the tree search is probabilistic, we repeated the analyses several times. Fig.~\ref{calibration} shows the results of 7 batches. 
A low number of visits gives a rather different policy, since it takes a few simulations for the Monte-Carlo algorithm to balance the exploitation/exploration ratio. After that we see an increasing KL-divergence value. Due to practical considerations, we chose 100,000 visits for further experiments.

We analyzed four full games with respect to the hit rates of four different networks (Table \ref{HitRates}).
The games are chosen to be different in style and strength. The first is a historical game from 1846, the famous `ear-reddening' game \cite{power1998invincible}. The second game is from the Alphago vs.~Lee Sedol match in 2016. The second game of the match contains the famous move 37, an example that a computer can also have creative ideas. The third is taken from the 44th Meijin title match, as an example of post-AG professional play. And the fourth is an amateur game.
The results show the tendency of higher hit rates for stronger networks, both in terms of structure and length of training. Interestingly, the amateur game has the opposite tendency.

The percentages in Table \ref{HitRates} are reminiscent of the success rate of supervised learning for predicting human expert moves used in the first version of AG \cite{AlphaGo2016}. In the self-play based reinforcement learning the network is trying to predict the outcome of the tree search indirectly. So one might wonder whether it would be possible to improve the networks without any more self-play games; after all, the tree search is a short-circuited self-play. Of course, this could only work for fine-tuning of networks that are already strong, since the external reward signal is not available. We invite the deep learning community to test this hypothesis.

The above analysis has the problem that moves in a game are correlated. Therefore, we also measured KL-divergence over 915 games from the KGS server. All the games are between players of 4 dan or better, so they represent strong amateur play. We picked a random game from each and did a 100,000 visit analysis. Fig.~\ref{KGS} compares the KL-divergence in the early 20 block and the late 40 block Katago networks. We can observe that the stronger network has smaller KL-divergence values on average; also, the maximal values are more extreme.

\section{Application: Cheat Detection}

\begin{figure}
\includegraphics[width=0.5\textwidth]{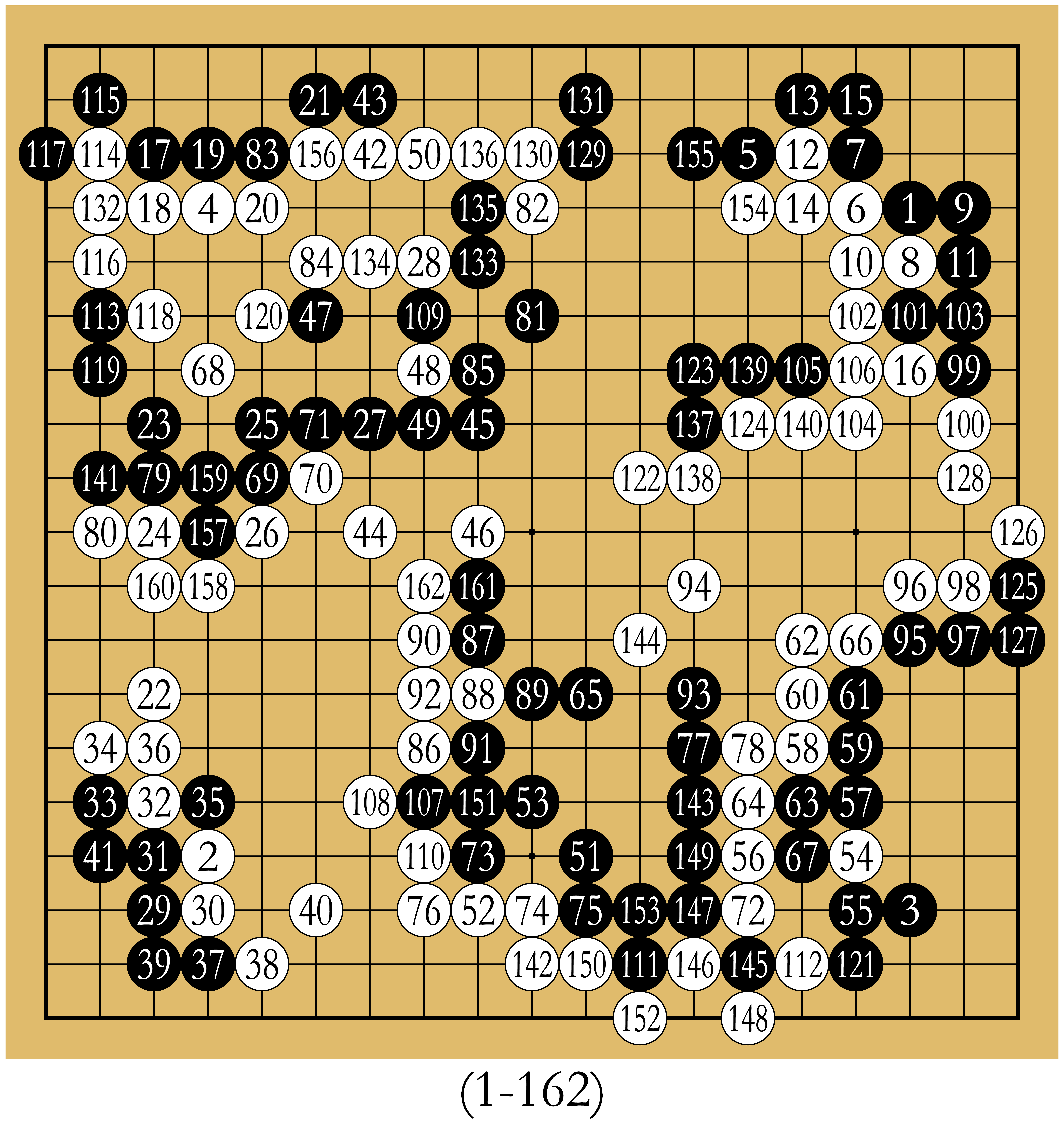}
\caption{Game 1: White is likely using an AI. The SGF files for the presented games are available upon request.}
\label{game1-figure}
\end{figure}

\begin{figure}
  \includegraphics[width=0.24\textwidth]{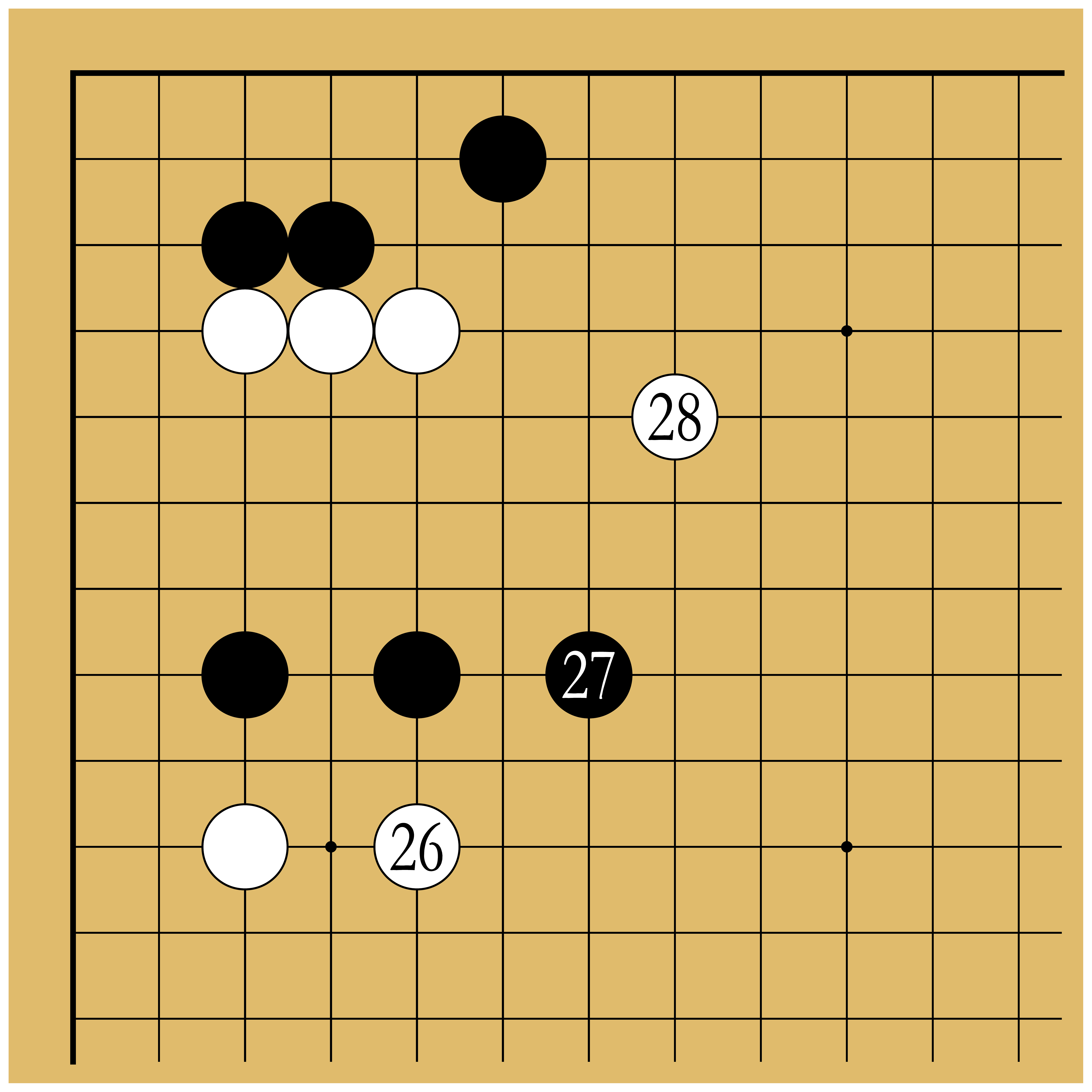}
  \includegraphics[width=0.24\textwidth]{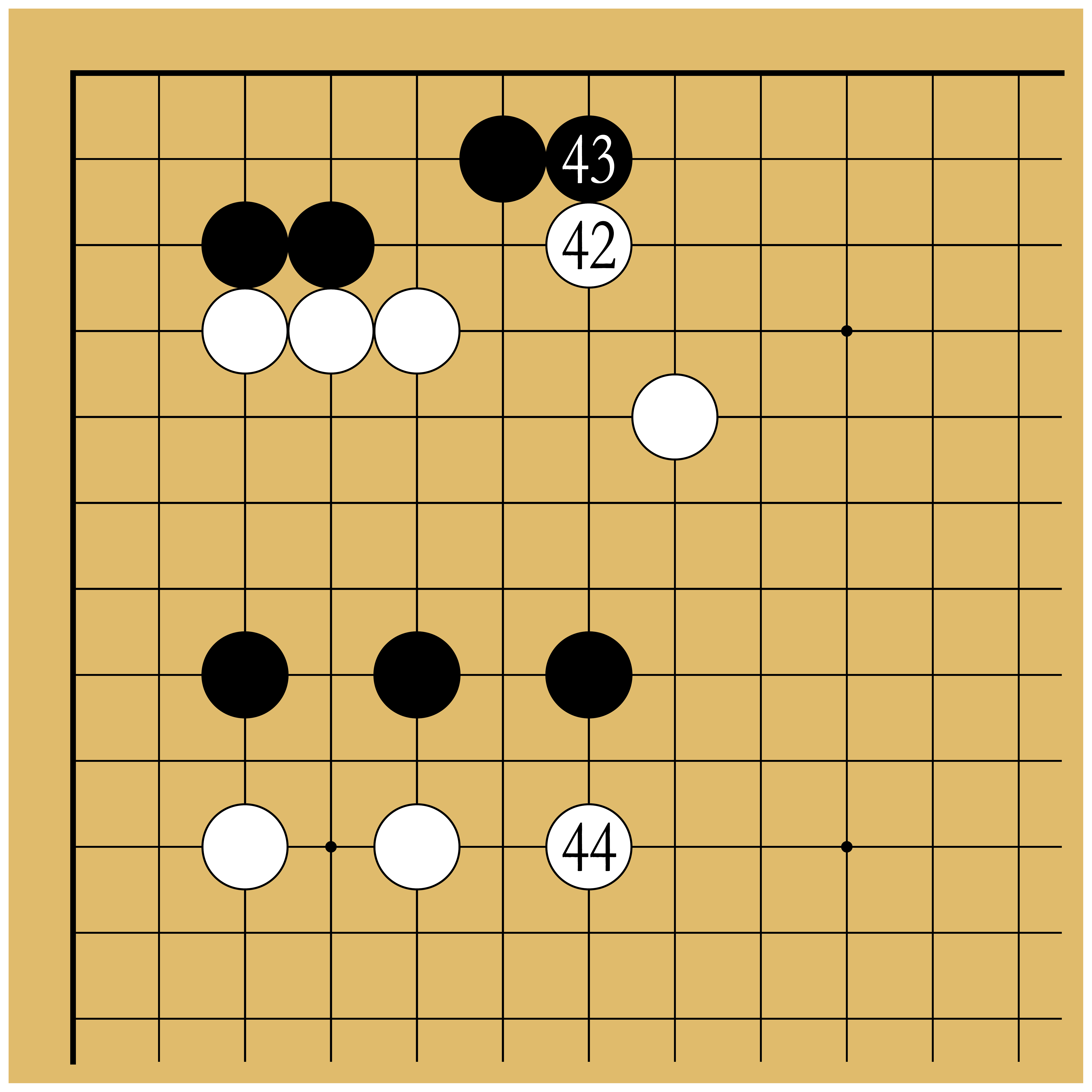}
  \caption{White 26, 28, 42, and 44 are exactly correct according to KataGo – even though a human player could think of many viable plans in these parts of the game.}
  \label{game1-sample}
  \end{figure}

\begin{figure*}[h]
  \includegraphics[width=\textwidth]{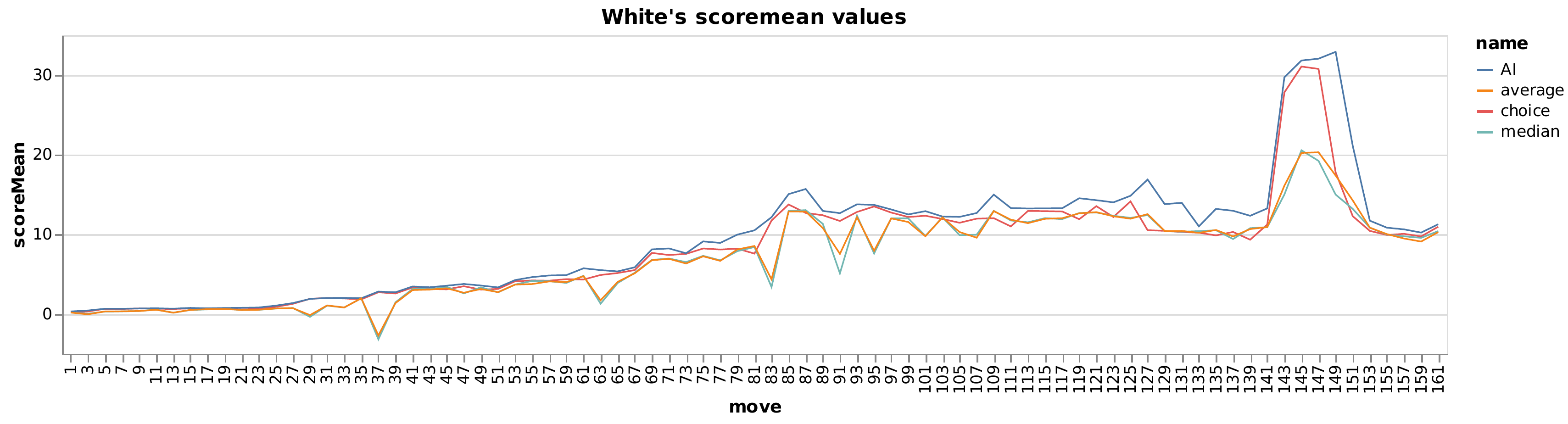}
    \includegraphics[width=\textwidth]{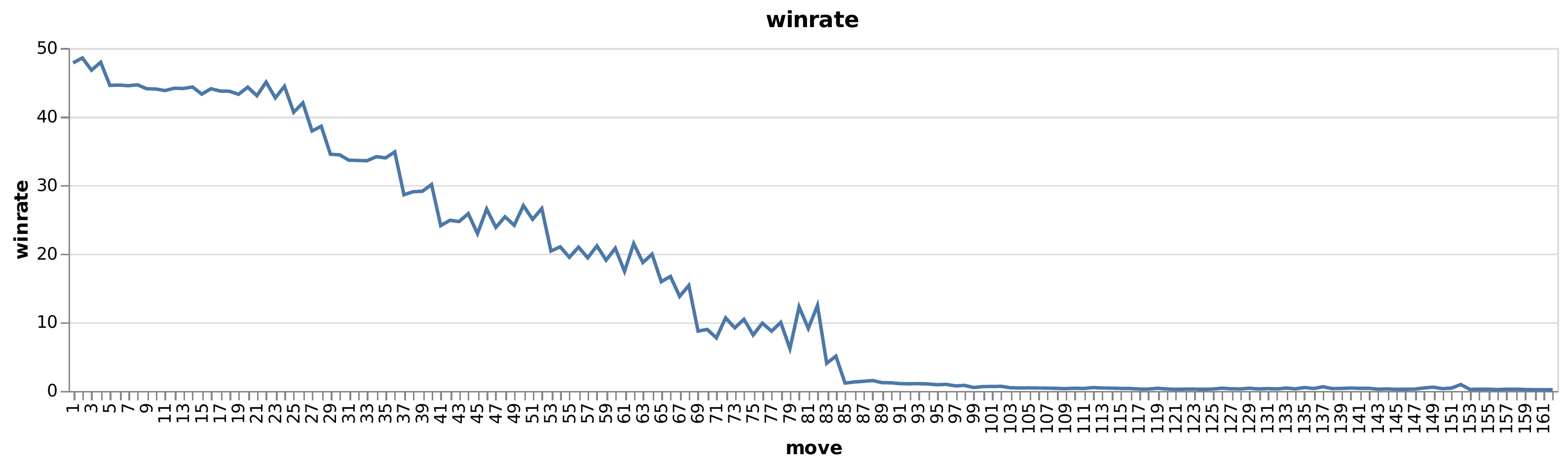}
    \caption{White's score mean and win rate graphs for Game 1. In order to indicate the nature of the board position (whether only a single `forced' move is available, or there are several equally good options), we display the AI's best move, the actual choice made by the player, and the average and median of the candidate moves considered by the engine.
      Before move 86, when White’s win rate hits 98\%, his moves were almost perfect. Afterwards, White’s play becomes less sharp, as indicated by the distance of the AI and choice lines; but the win rate does not change, suggesting an AI’s ‘safe play mode’.}
\label{game1-graphs}
  \end{figure*}

 Using an AI engine for finding best moves and variations in a game is called
\emph{analysis} after the game is finished; and
\emph{cheating} when the game is still ongoing.

The widespread availability of AI engines is beneficial in many ways. Most notably, one can improve their playing skills by reviewing their games with an AI. However, there are downsides of the technological progress: many players report cheating on online Go servers. With the availability of superhuman AI engines, online cheating might be rampant; but, besides the rare cases where a player admitted to cheating, there is no direct evidence for this except the gut feelings of strong human players.

Cheating defeats the purpose of online playing, where one wants to have a human opponent. On Asian servers, the top ranks are reportedly infested by cheaters. This has resulted in previously top-ranked humans to drop to lower ranks, starting a snowball effect inside the servers’ ranking systems. If no countermeasures to cheating are found, in the near future it is possible that online ratings will be largely devalued.

Strong players can quickly and reliably assess the opponent's strength. Consequently, experienced players can recognize superhuman AI opponents. Could this be reproduced or at least helped by software tools?

Players with a rating history are easier to catch from cheating by noticing a sudden increase in their won games. However, clever cheaters that only consult an AI occasionally may be impossible to detect this way. Also, the availability of AI-based training tools may accelerate individual learning.

In this research, we do not consider players' histories, so we can deal with newly registered users as well. Our aim is therefore to be able to decide whether cheating happened in a single game solely based on the game record.



\subsubsection*{Prior work}

Chess has a longer history of living with superhuman AI engines, thus the integrity of online games has been investigated extensively. However, the conclusion is that fully automated cheat detection is not possible. In \cite{BARNES2015} it is demonstrated that `false positives' are abundant. This was shown by the existence of historic games that would be classified as cheating, though that clearly could not have happened.

 In \cite{Coquid2017}, the theory of complex networks and the PageRank
algorithm was used to find distinguishing statistical features of
human and computer play. The analysis was based on local information
(3$\times$3 squares) and did not use the modern capabilities of
superhuman AIs.

\subsection{Human ways of recognizing an AI-using cheater}

 The ways that human players recognize AI-using cheaters, listed in this section, might be of help in designing software tools for automatically catching cheaters.

\subsubsection{Temporal evidence}

 When a cheater consults an AI, there is a near-constant time lag created by the cheater inputting their opponent’s last move to the AI program and waiting a moment for the AI to come up with an answer. When done in a straightforward fashion, this results in a cheater always playing their move after for example five seconds – no matter if the move is obvious or extremely difficult for a human player to come up with.

\subsubsection{Playing style}

It did not take long for human players to notice that AI engines have a discernible playing style, emphasising quick exchanges and maintaining a whole-board balance. At first the difference to human players was glaring, but human players have since adopted the AI’s favored techniques, resulting in a human-AI blend.

Still, there are many moments during games when, according to the AI, an ‘obvious’ move by human intuition is wrong, with the correct move being something very unintuitive. When several such moves get played by the same player in a single game, the player is suspicious.

\subsubsection{Safe play when ahead}

 As most AI engines choose their moves by the win rate estimate, when a game is deemed practically ‘over’ (at roughly 98\% and above), they will start playing moves that are not optimal in terms of the score mean but that still retain the player’s win rate. This leads to the AI choosing moves that a human player would consider ‘slack’, and often a strong human player can notice when their opponent enters this kind of ‘safe play mode’.

\subsubsection{Seemingly inconsistent play}

 Human players and AI engines choose their moves very differently. Strong human players generally:

\begin{enumerate}
\item analyse and judge the current whole-board situation,
\item try to identify the most important or valuable areas of the board,
\item create a plan for how to develop the game, and
\item finally choose a move that furthers the plan.
\end{enumerate}

This process is then more or less repeated on each move, with adjustments made as necessary depending on what the opponent is doing.

As the opponent generally acts on a similar modus operandi, it becomes valuable for a strong player to try to infer what the opponent is planning and to adjust their own plan accordingly. For strong human players, this generates a kind of non-verbal discussion or give-and-take that takes place on the go board. For this reason, Go is sometimes referred to as ‘hand talk’ in Asian countries.

As the AI does not form plans in a similar way as humans, it is not possible for a human to create this kind of a higher-level discussion with an AI engine. The AI will constantly play moves that, to a human, seem to betray its plan – possibly only because the human player is unable to grasp it.

\subsection{Case studies}

In this section, we have analysed four games, three of which (most likely) involve cheaters. All four games were played online and analysed by a professional Go player. As we have no input from the other player, ultimately there is no hard evidence on whether they were cheating or not.

The difficulty of identifying a cheater depends greatly on whether the cheater is trying to cover their cheating or not. A clever cheater will vary the time they use for their moves, playing ‘obvious’ moves quickly and taking more time for difficult moves; and they will also not always play the AI’s best recommended move. Additionally, as AI engines rarely make big mistakes (especially early on in the game), a clever cheater would optimally try to include a few larger mistakes in their play.

The analysis has been performed as follows: first, the win rate graph of the game is checked. Since a cheater is using the AI to win the game, the win rate graph will generally tend to be one-sided, steadily rising to 99\%; large shifts should not take place, as even a strong AI engine might not be able to beat a strong human if it falls too much behind. An exception is if both players are cheating, in which case the win rate usually progresses evenly for the most of the game.

Secondly, the development of the player's average effect during the game is checked. Of particular interest are the final average effect for the whole game, which is a general indicator of the player's skill, and if the players' average effects develop in similar stages. Also, a player's moves after their win rate reached 98\% can be indicative of AI involvement, as an AI will start playing score mean inefficient moves after this point.

Thirdly, we check how the player performed in comparison to KataGo's move recommendations. If the player played moves that are roughly as good as KataGo's first recommendations, the player is suspect; whereas, if the player does considerably worse than KataGo, that is evidence of either human play or at least the player avoiding the AI's best recommended moves.

\subsection{Game 1}

The white player in Fig.~\ref{game1-figure} is most likely consulting an AI.
Firstly, an experienced human player can already find White’s opening suspicious when comparing White’s choices with the AI’s suggestions. 26 and 28, shown in Fig.~\ref{game1-sample}, are non-obvious moves to a human but first options for the AI. A bit later, 42 and 44 are another combination that looks made-up on the go, but exactly matches the AI’s recommendation. For a third example, 58 and its follow-up are very rarely seen in human play and, while not KataGo’s first recommendation, perform just about as well.

Secondly, as shown in Fig.~\ref{game1-graphs} White basically makes no mistakes up until 86, even though Black is a professional player. This is difficult to accomplish even for a top human player.

Thirdly, after White reaches 98\% win rate at move 86 as shown in Fig.~\ref{game1-graphs}, White’s play gets sloppy in terms of the score mean. After this point, the white average effect starts decreasing, but the win rate is firmly stuck at 99\%.

All three pieces of evidence put together, it is very likely that an AI engine was involved.

\subsection{Game 2}
\begin{figure}
    \includegraphics[width=0.5\textwidth]{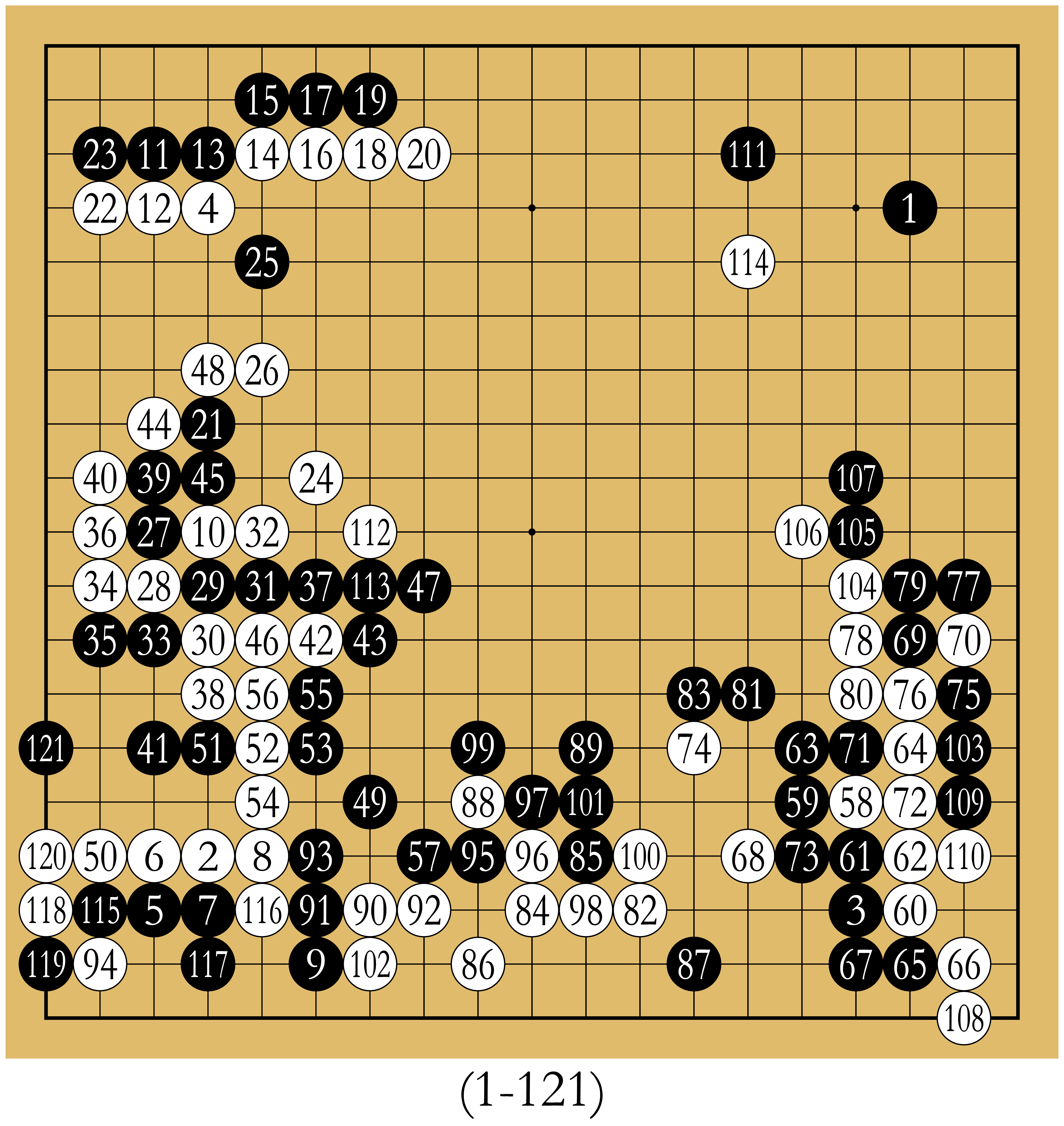}
\caption{Game 2: Both players are likely using an AI.}
\label{game2-figure}
\end{figure}
\begin{figure*}
    \includegraphics[width=0.95\textwidth]{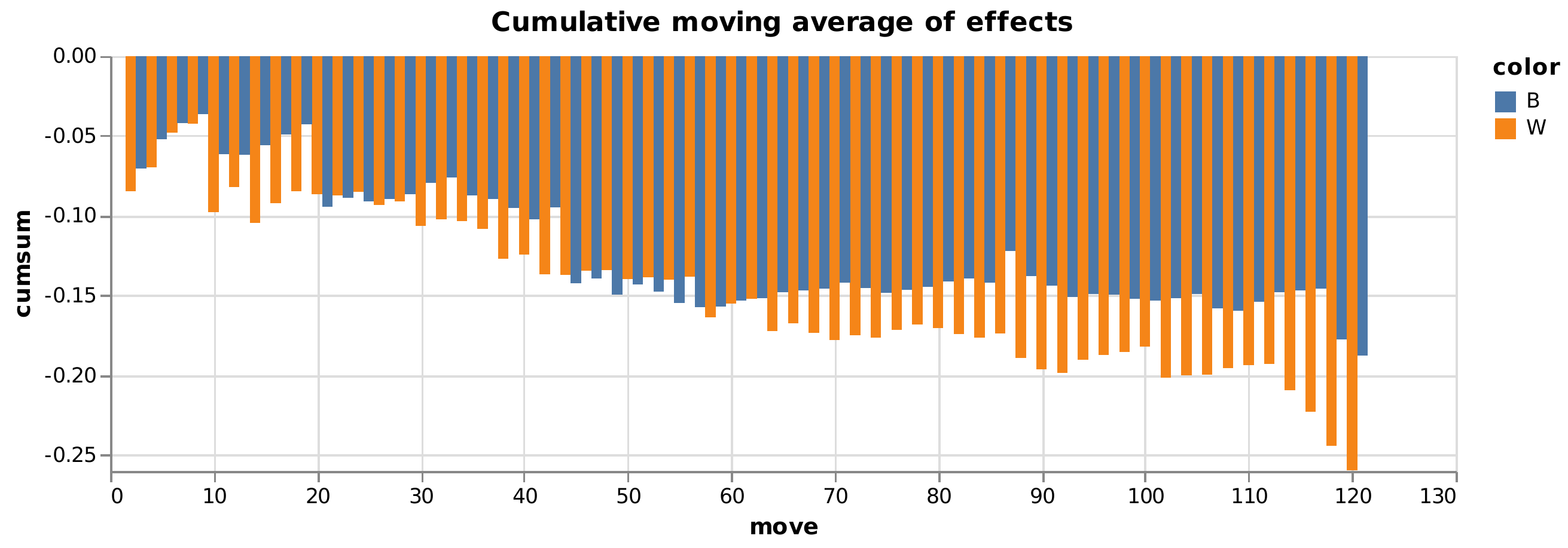}
  \includegraphics[width=0.95\textwidth]{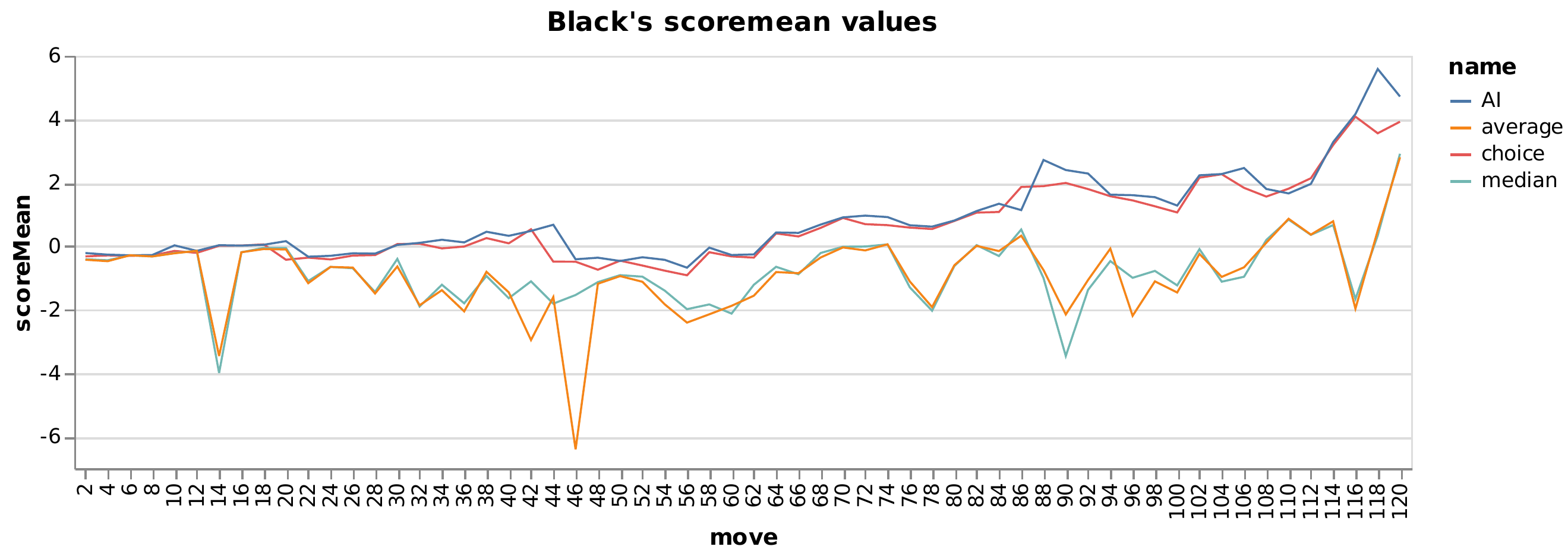}
  \includegraphics[width=0.95\textwidth]{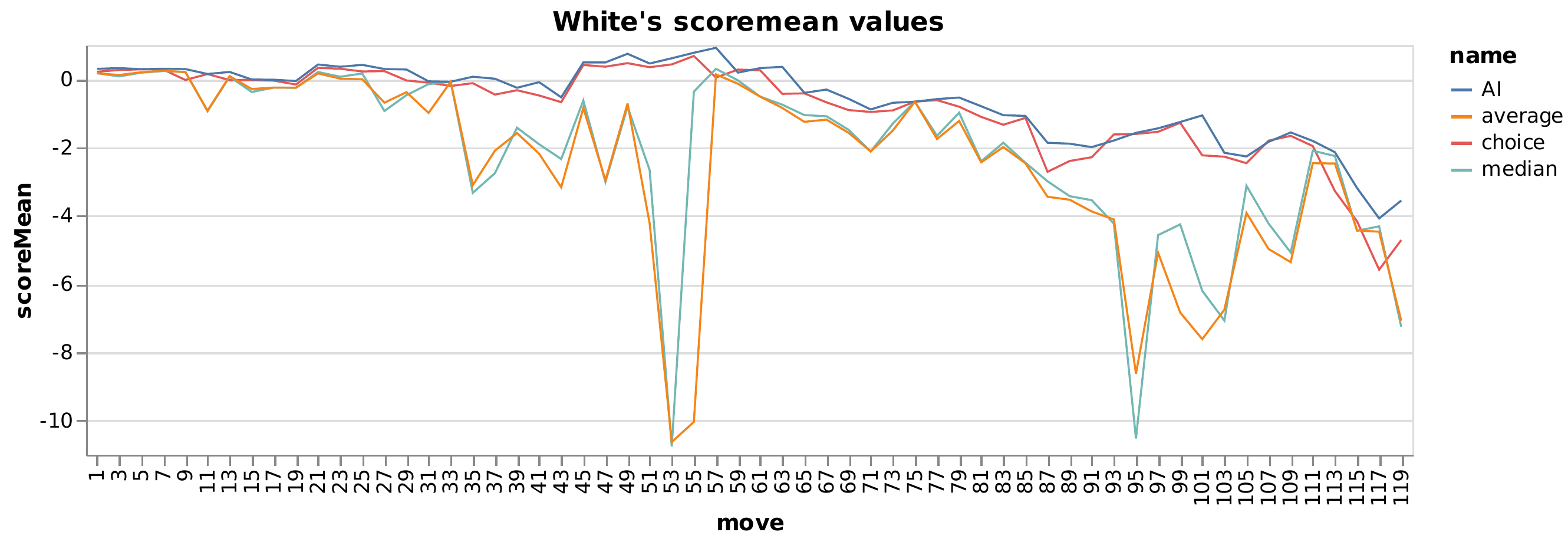}
\caption{The two players’ average effects and score means for Game 2. Both players’ average effects are considerably small when taking into amount the ‘volatility’ of the game, indicated by the distance of the AI, average, and median lines in the score mean graphs.}
\label{game2-graphs}
  \end{figure*}

Both players in Fig.~\ref{game2-figure} are most likely consulting an AI.
Most of the moves in this game are among KataGo’s top picks. Furthermore, the players’ average effects are extremely small ($-0.25$ and $-0.20$) even though there is a large variance in the score means of KataGo’s considered moves, as shown in Fig.~\ref{game2-graphs}. Even the world champion of Go would find it difficult to play this well.

\begin{figure}
\includegraphics[width=0.5\textwidth]{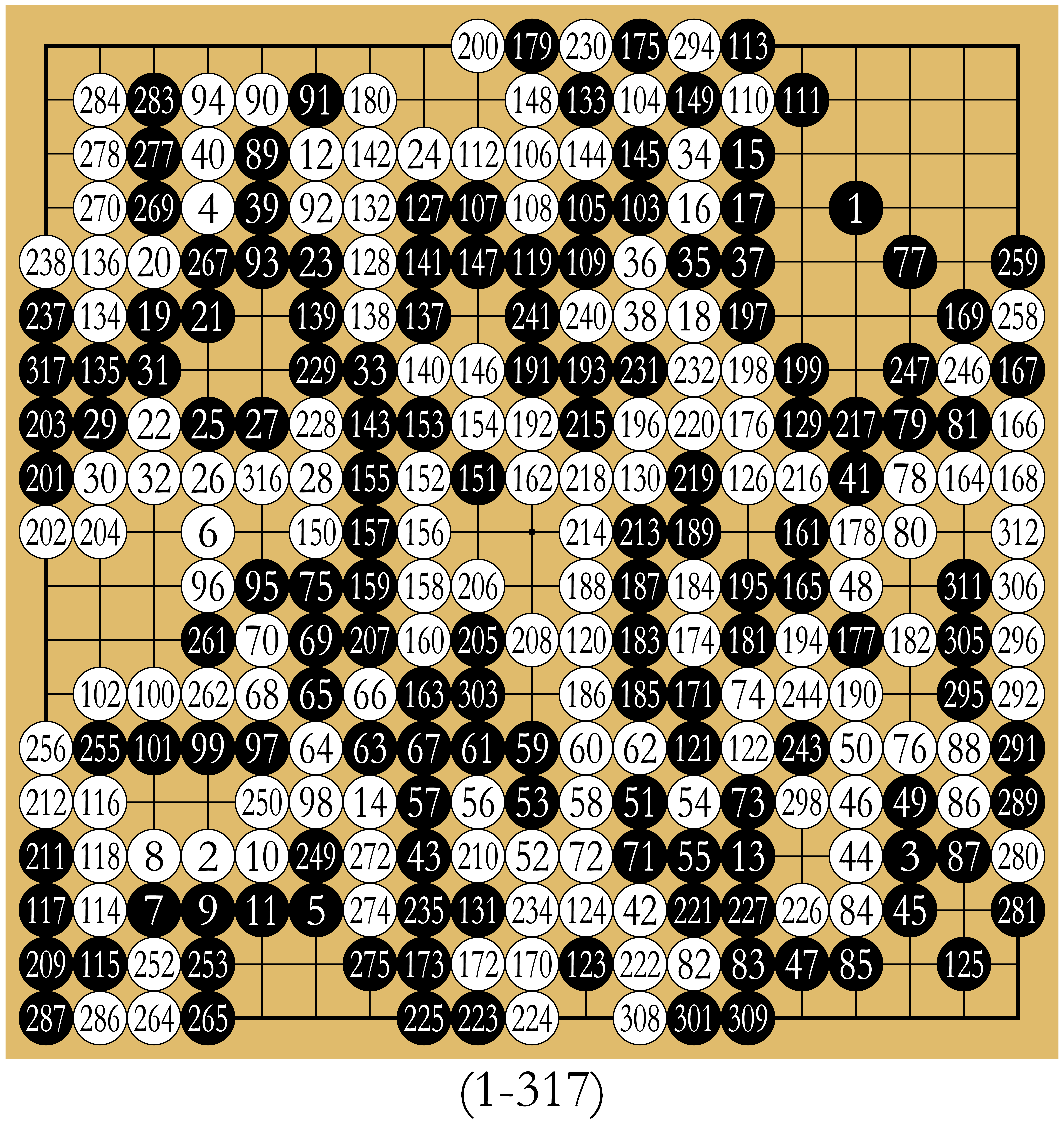}
\caption{Game 3: Neither player is likely using an AI.}
\label{game3-figure}
\end{figure}
\begin{figure*}
    \includegraphics[width=\textwidth]{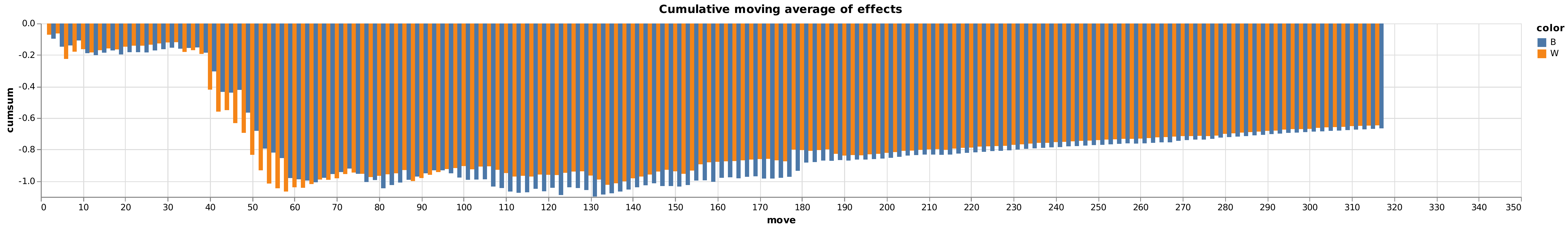}
  \includegraphics[width=\textwidth]{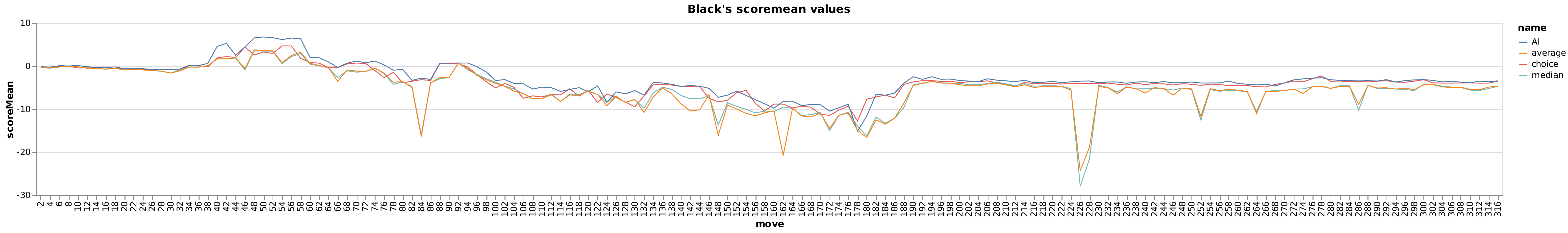}
  \includegraphics[width=\textwidth]{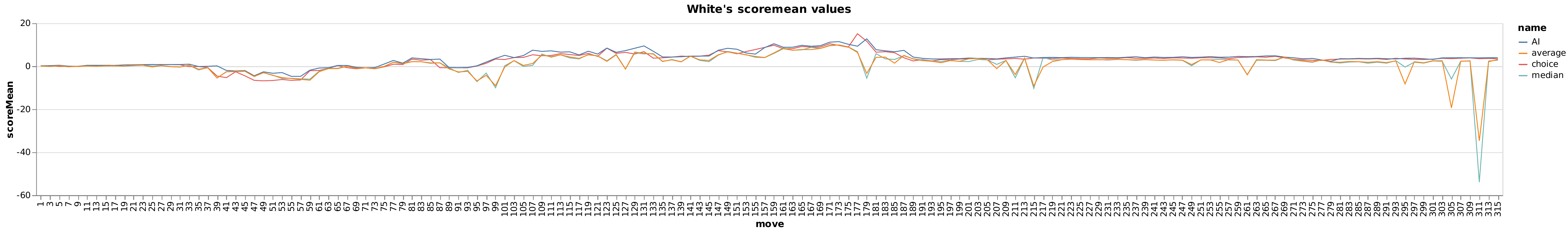}
  \includegraphics[width=\textwidth]{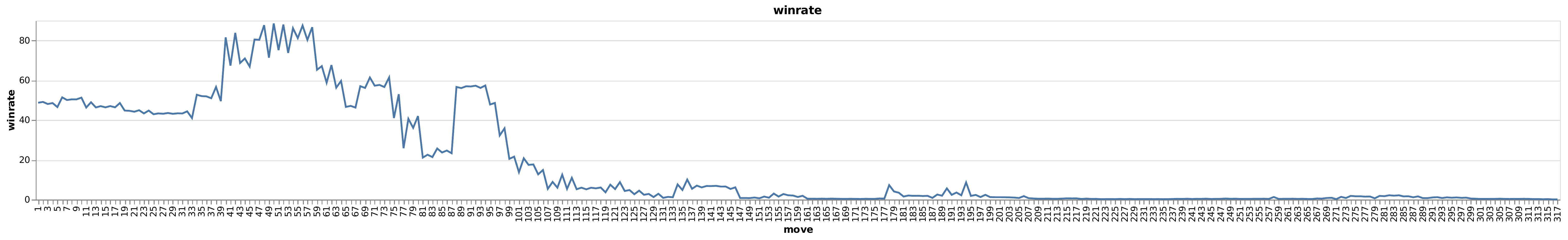}
\caption{The two players’ average effects, score means, and the win rate for Game 3. The fairly low size of the players’ average effects, the variance in the score mean graphs, and the up-and-down in the win rate graph suggest that this was a game by strong human players.}
\label{game3-graphs}
  \end{figure*}

\subsection{Game 3}

 Most likely neither player in Fig.~\ref{game3-figure} consulted an AI – this is a game by strong human players.
As shown in Fig.~\ref{game3-graphs}, the average effect for the players peaks at around $-1.0$ and finally settles to around $-0.65$ for each, which are reasonable numbers for strong human players. Comparing the players’ chosen moves with KataGo’s recommended alternatives, we see that both players generally perform better than the average choice but worse than the best choice, with plenty of exceptions to both directions.

An AI-using smart cheater might attempt to play bad moves from time to time, but not so much that it should threaten their win. The win rate graph in Fig.~\ref{game3-graphs} shows that this is not the case, as there are large shifts in the win rate in the first third of the game: first Black got a considerable lead, then White turned the game around, after which Black caught up again, after which White took off to a decisive lead. For further evidence, White’s win rate wavers even after first hitting 99\%, which is common to human games.

While it is impossible to prove that neither player used the AI at any point during the game, it does not look like an AI was consulted to decide the outcome of the game.

\begin{figure}
\includegraphics[width=0.5\textwidth]{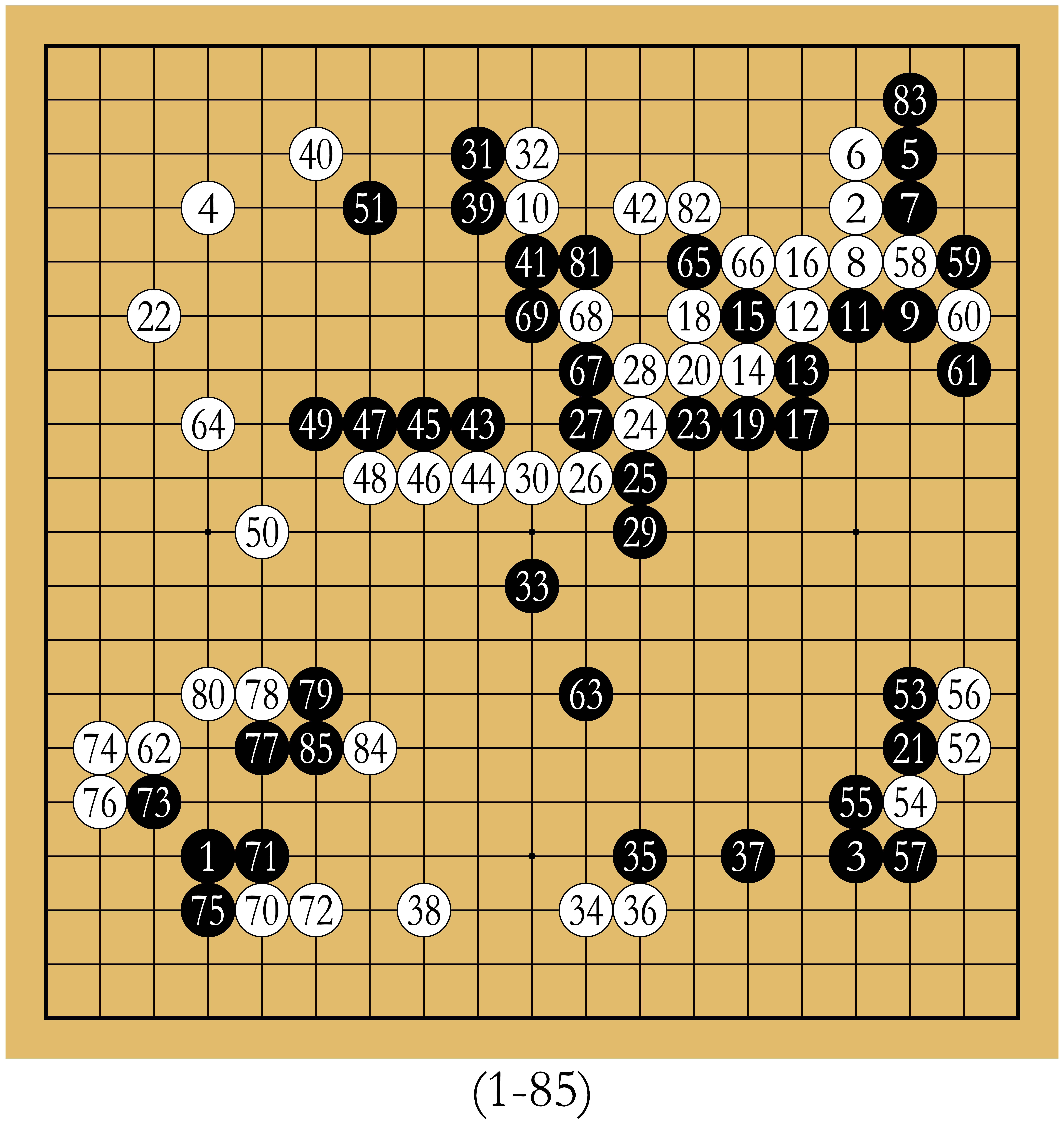}
\caption{Game 4: Black is likely using an AI.}
\label{game4-figure}
\end{figure}
\begin{figure}
  \includegraphics[width=0.24\textwidth]{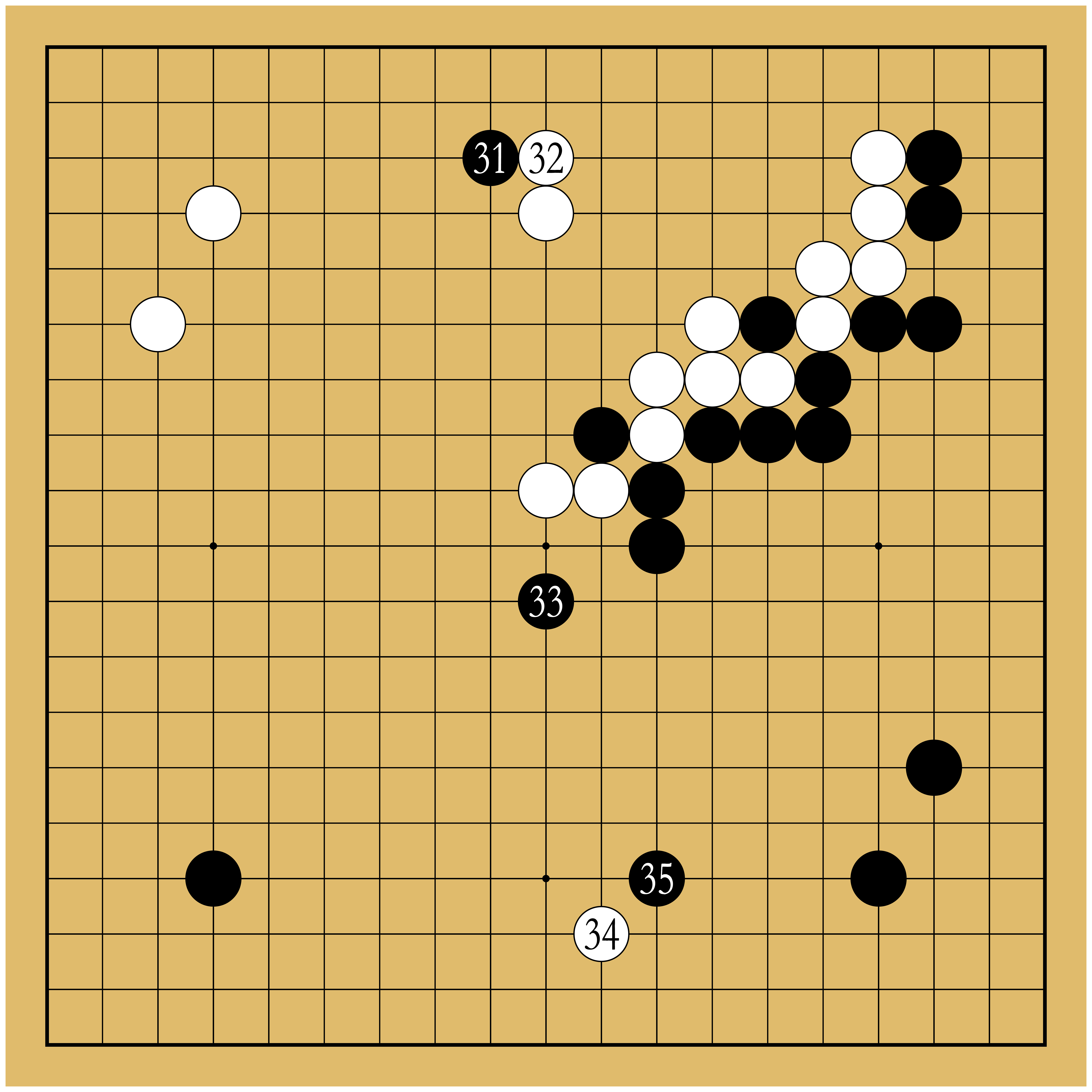}
  \includegraphics[width=0.24\textwidth]{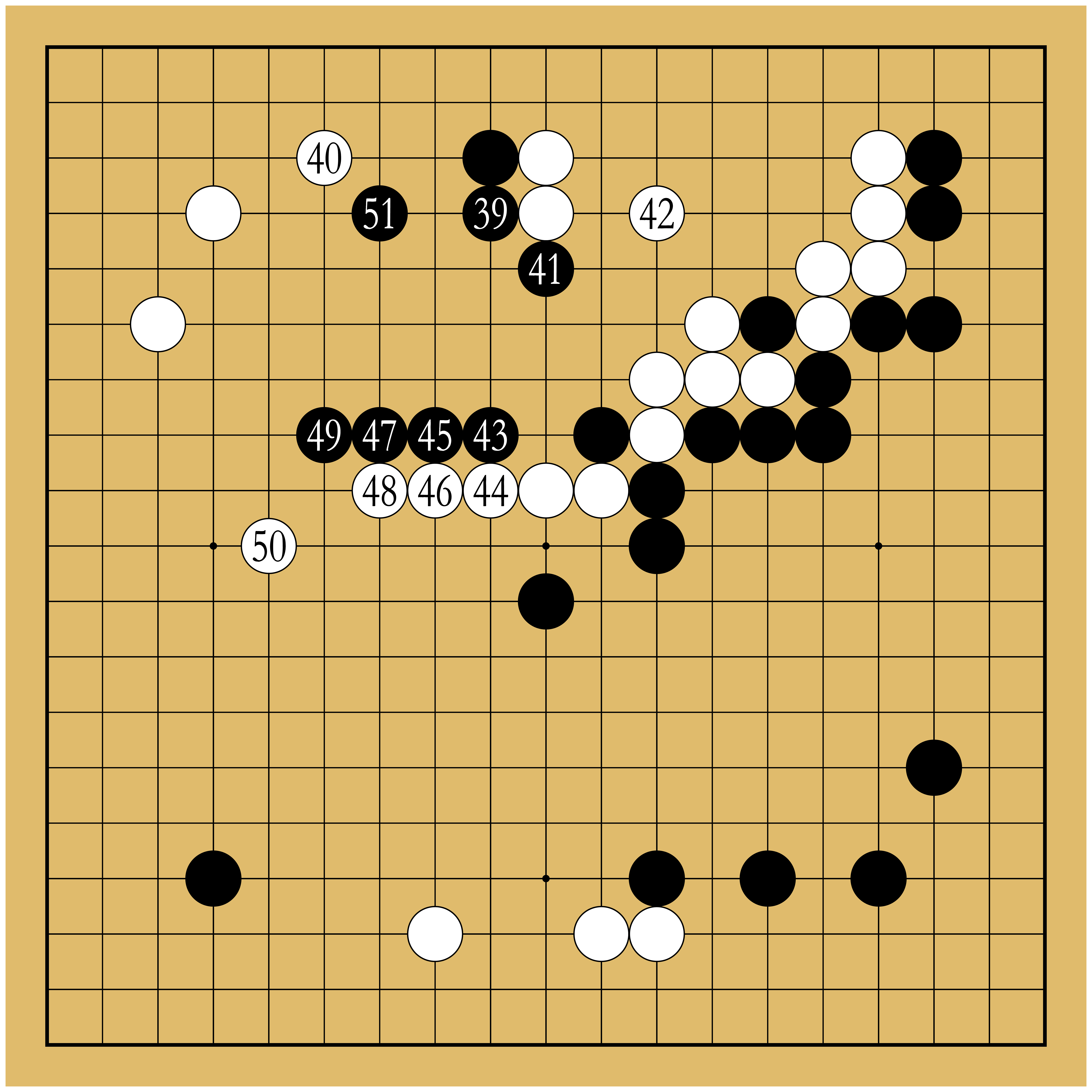}
  \caption{Most of these black moves would be difficult for a human to come up with, but they align with KataGo's recommendations.}
  \label{game4-sample}
  \end{figure}
\begin{figure*}
  \includegraphics[width=0.5\textwidth]{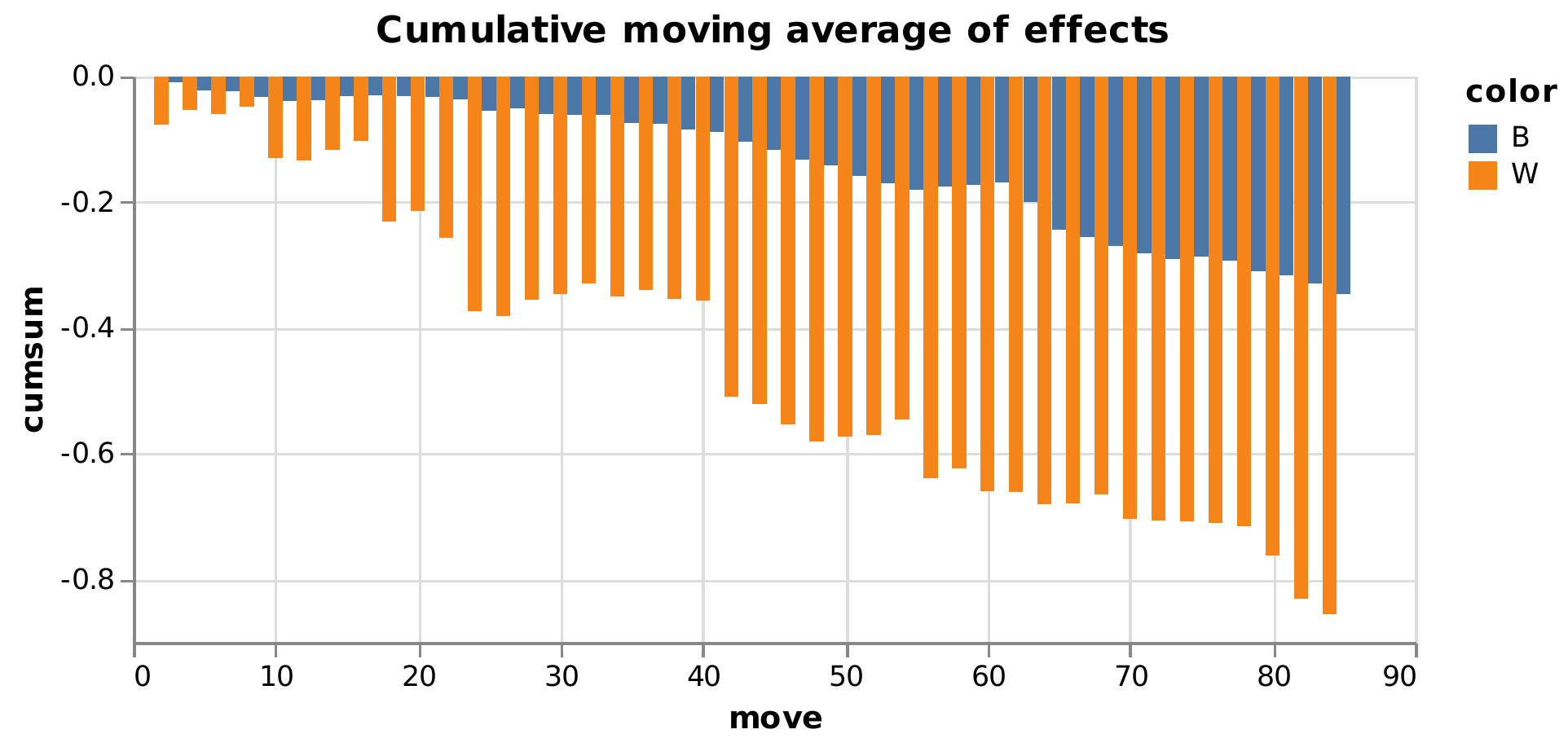}
  \includegraphics[width=0.5\textwidth]{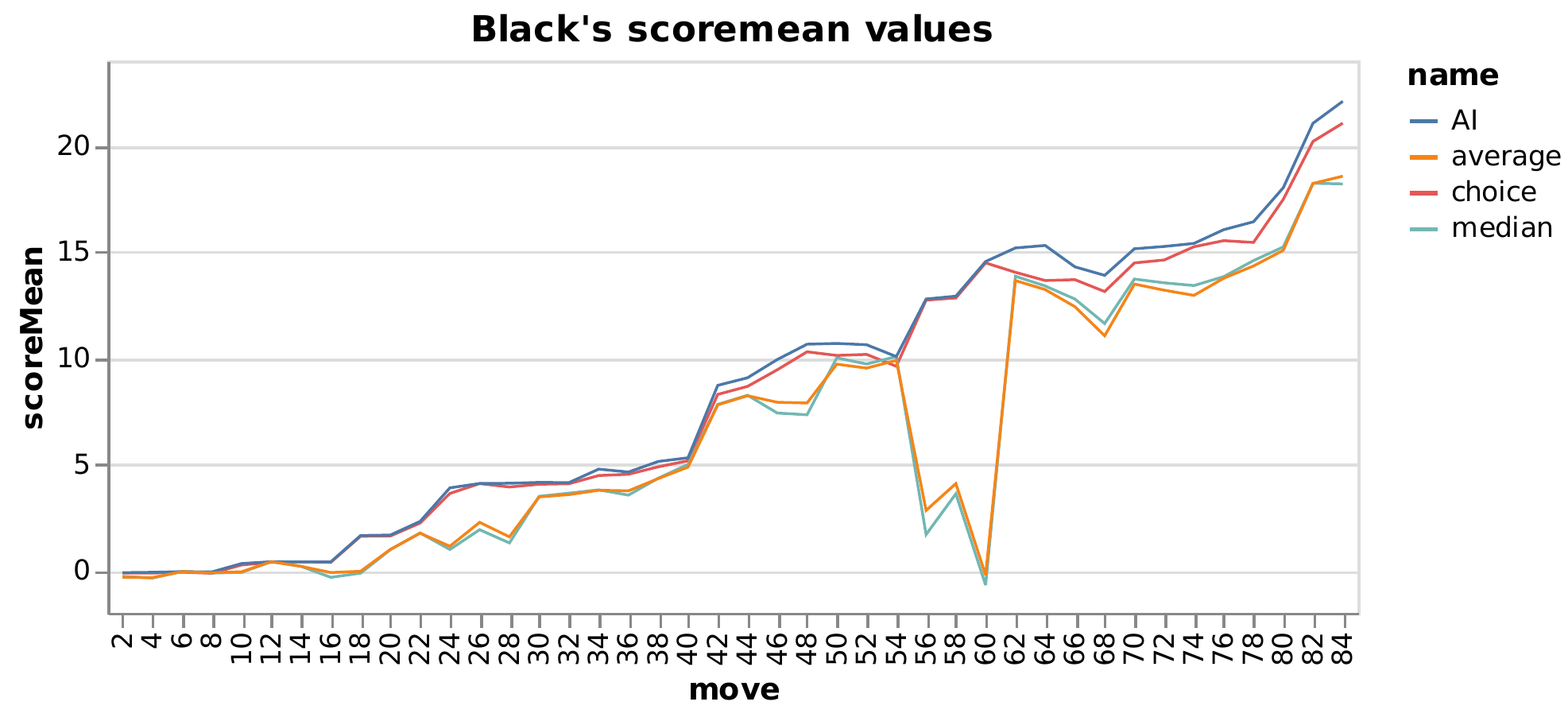}
  \includegraphics[width=0.5\textwidth]{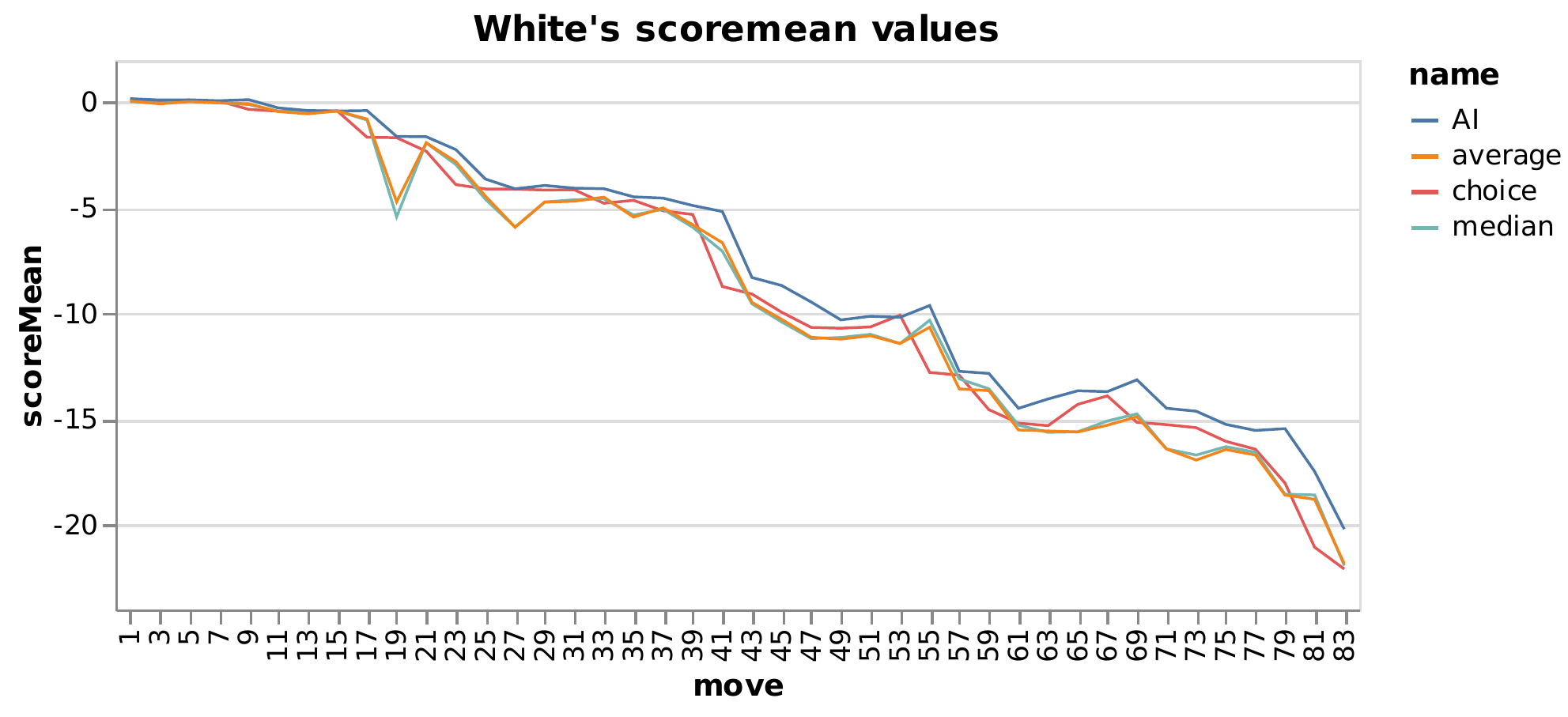}
  \includegraphics[width=0.5\textwidth]{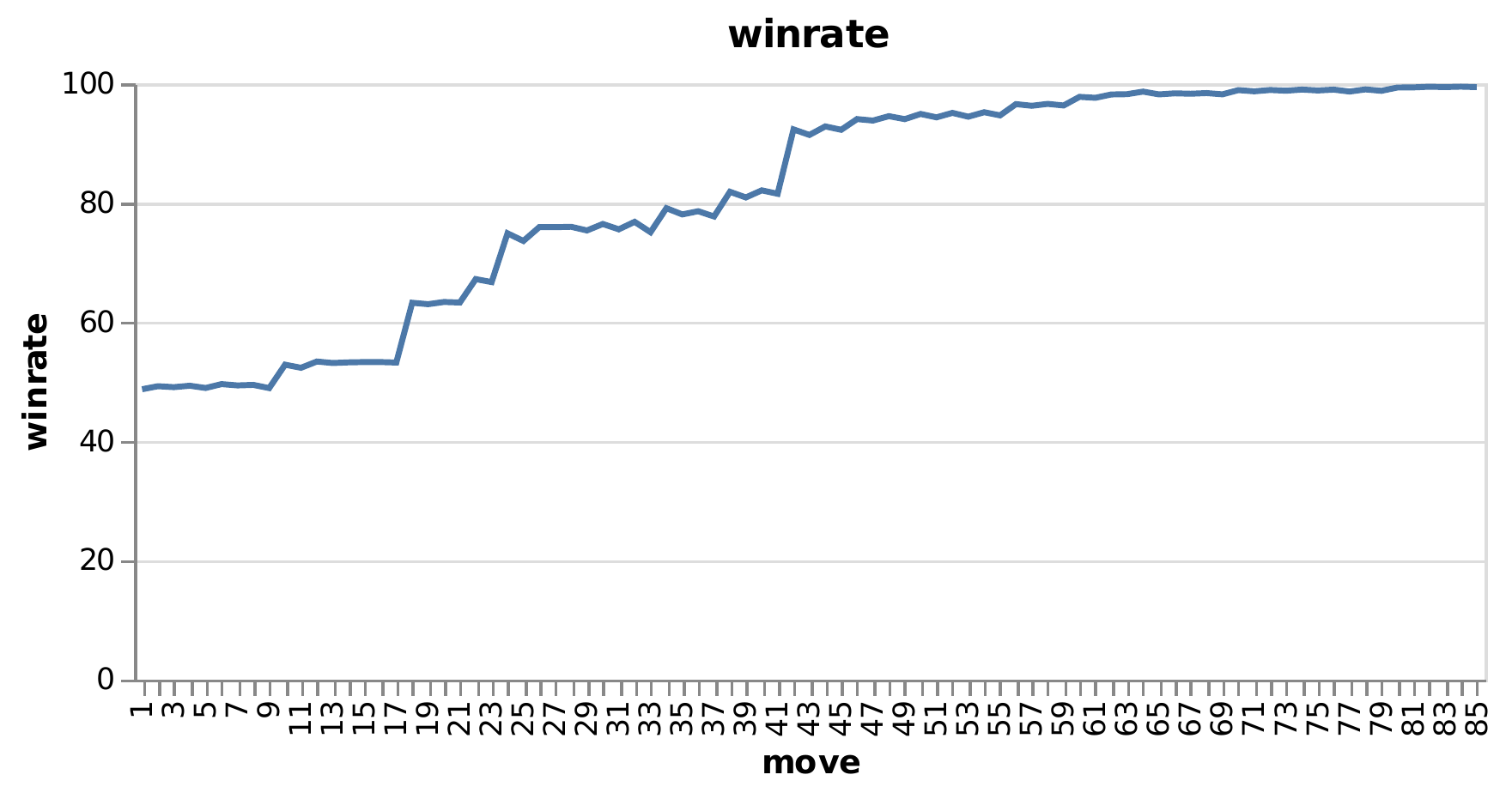}
\caption{Black’s average effect, both players' score means, and the win rate for Game 4. The straightforwardness of the win rate graph as well as Black’s small average effect suggest AI involvement.}
\label{game4-graphs}
  \end{figure*}

\subsection{Game 4}

The black player in Fig.~\ref{game4-figure} is most likely consulting an AI.
This case is possibly the most obvious to a strong human player. First, black 31 and 35 in Fig.~\ref{game4-sample} are moves that few human players could consider. Then, Black’s play from 39 to 51, after which Black lives comfortably in the centre, would also be unthinkable to most – but all of these black moves are KataGo’s first recommendations. A bit later, black 63 also looks mistimed in human terms, but is among KataGo’s top choices.

Secondly, looking at the win rate graph in Fig.~\ref{game4-graphs}, Black’s win rate is headed directly to 99\% with practically no drops. This is evidence of a vast difference of skill between the players – even though White is a professional player who did not play particularly badly in this game, according to KataGo.

Thirdly, looking at the size of Black’s average effect in Fig.~\ref{game4-graphs}, we see that Black manages an impressive $-0.16$ until move 61, at which point Black’s win rate has reached 98\%. After this, Black’s moves get sloppier in terms of the score mean, which further suggests an AI.

All three pieces of evidence put together, it is very likely that an AI engine was involved.

\section{Software Implementation}

We developed a dedicated software package for the described computations and for generating the diagrams. The source code of LambdaGo is available at \url {https://github.com/egri-nagy/lambdago}.
It is a simple command-line tool.

The core system (including a game engine) is written in the Clojure language \url{https://www.clojure.org}. Due to its dynamic nature, this functional language is particularly suited for data-driven experimentation \cite{Clojure2020}. It is hosted on the JVM, therefore it also has convenient access to the whole JAVA ecosystem.

For parsing the game record SGF files \cite{SGF}, in order to avoid writing yet another parser, we use a parser generator, Instaparse \url{https://github.com/Engelberg/instaparse}. This library is based on the idea of parsing with derivatives \cite{MattMight2011}.

The visualization of the graphs is done by the Vega-lite library \cite{2017-vega-lite}. It is a high-level grammar of graphics that allowed us automate the task of diagram generation.
The diagrams are just data (JSON files), which can be manipulated easily in LambdaGo before rendering in a browser.
The Go diagrams are made with GOWrite 2 \cite{gowrite}, a high-quality Go publishing tool.

The workflow of the system evolved through the cheat-detection application, and it has two steps: analysis and visualization.

\subsubsection*{Analysis} The analysis can be done by the Lizzie GUI application (\url{https://github.com/featurecat/lizzie}). This was designed as an interface to Leela Zero \cite{LZ}, but later it was adapted to work with other engines as well. It produces SGF files with the analysis information added. The KataGo engine \cite{KataGo2019} also has a direct interface to its analysis engine, which accepts and emits information in the ubiquitous JSON format. The analysis is a GPU-intensive and time consuming  computation, so for practical reasons we need to limit the visit counts.
\subsubsection*{Visualization} The output of the analysis can be quickly processed to generate the diagrams. They can be generated in batch mode as well.
We expect that these visualization features will appear in other tools as well, as the analysis needs of the users will reach more sophisticated levels.

\section{Conclusion}

Building upon the advances in artificial intelligence, and the developments in open-source software projects, we suggested novel measures for evaluating and understanding AI game analyses. Measuring the search gap (the added value of the tree search to the raw output of the neural network) allows us to measure the strength of the network intrinsically, without playing other networks. The effect of a move can be used for assessing a player's performance with high resolution (move by move). We showed that an investigation of the effect can be helpful in detecting online cheating. Although automated cheat-detection may never be feasible due to the danger of false positives, we used these tools in a real online tournament and could catch a cheating player, who admitted the misconduct. This is an example of a successful collaboration of a human arbiter and an AI engine, according to the human-plus-machine paradigm envisioned by former chess world champion Garry Kasparov\cite{kasparov2017deep}. What happens in the world of the game of Go will happen in other aspects of our life, and therefore it is valuable to understand the effects of AI technologies on the game.

In this paper we demonstrated that deriving further measures based on the inner parameters and outputs of deep learning AI engines could provide useful tools for solving practical problems and ways to advance theoretical Go knowledge.

\vskip1ex

\noindent\textbf{Acknowledgment.} This paper benefited from conversations with Go AI tool developers Sander Land (KaTrain), Benjamin Teuber (AI Sensei) and David Wu (KataGo).

\bibliographystyle{plain}
\bibliography{../compgo}

\end{document}